\documentclass{ifacconf}

\usepackage{import} 
\usepackage{graphicx} 
\usepackage{natbib}           
\usepackage{amsmath}
\usepackage{amssymb}
\usepackage{mathtools} 
\usepackage{bm} 

\usepackage{grffile} 
\usepackage[tight,footnotesize]{subfigure}
\usepackage{microtype} 
\usepackage{url}

\usepackage{verbatim} 
\usepackage{comment}
\usepackage{arydshln}

\usepackage{enumitem}

\usepackage{adjustbox}
\usepackage{pgfplots}

\usepackage{etoolbox}
\robustify{\cite}

\newenvironment{lemma}{\begin{lem}}{\end{lem}}

\newenvironment{proof}{\begin{pf}}{\end{pf}}

\newenvironment{remark}{\begin{rem}}{\end{rem}}
\newenvironment{problem}{\begin{prob}}{\end{prob}}

\newcommand{\cmmnt}[1]{}

\definecolor{RED}{RGB}{255,0,0}
\definecolor{BLUE}{RGB}{0,0,255}

\begin{document}
\begin{frontmatter}

\title{Distributed 3D Source Seeking via \text{SO}(3) Geometric Control of Robot Swarms} 

\thanks[footnoteinfo]{This work is supported by the ERC Starting Grant \emph{iSwarm} 101076091 and the RYC2020-030090-I grant from the Spanish Ministry of Science.}

\author[UGR]{Jesús Bautista} 
\author[UGR]{Héctor García de Marina} 

	\address[UGR]{Department of Computer Engineering, Automation \& Robotics, and Institute of Mathematics (IMAG) of University of Granada, Granada 18012, Spain \\(e-mails: \{jesusbv, hgdemarina\}@ugr.es).}

\begin{abstract} 
This paper presents a geometric control framework on the Lie group $\mathrm{SO}(3)$ for 3D source-seeking by robots with first-order attitude dynamics and constant translational speed. By working directly on $\mathrm{SO}(3)$, the approach avoids Euler-angle singularities and quaternion ambiguities, providing a unique, intrinsic representation of orientation. We design a proportional feed-forward controller that ensures exponential alignment of each agent to an estimated ascending direction toward a 3D scalar field source. The controller adapts to bounded unknown variations and preserves well-posed swarm formations. Numerical simulations demonstrate the effectiveness of the method, with all code provided open source for reproducibility.
\end{abstract}

\begin{keyword}
    Multi-agent systems, Robotics, Mobile Robots, Attitude Control, Nonlinear Control Systems
\end{keyword}

\end{frontmatter}


\section{Introduction} 
\label{sec: intro}

Detecting and following hydrothermal vents in the ocean to study geophysical processes \cite{german2008hydrothermal}, identifying the source of an electromagnetic field to assist search \& rescue operarions \cite{Liu2013}, tracking toxic gas plumes to provide early warning of volcanic eruptions \cite{Ilyinskaya2021} or industrial accidents, localizing radiation sources for rapid containment \cite{West2021} — these are all instances of a common robotics challenge: locating the maximum of an unknown scalar field. In such scenarios, robot swarms offer distinct advantages over single agents, as they can act as a flexible, distributed mesh of sensors that collectively adapt to environmental variations \cite{li2006moth}. By dispersing, coordinating, and sharing information, swarms can robustly estimate and track directions of increase in the field, thereby guiding the team toward the source even in the presence of disturbances, missing robots, or incomplete information.

At the algorithmic level, our recent works \cite{acuaviva2023resilient} and \cite{bautista2025resilient} introduced a fully distributed and resilient method that allows a swarm of robots to compute ascending directions autonomously, without relying on predefined formation shapes, and even in the presence of missing or misaligned individuals. While these works establish a strong foundation, realizing such strategies on practical robotic platforms requires additional control-level guarantees. Specifically, robots must not only compute the ascending direction in a distributed manner, but also (i) align their forward heading with this direction, (ii) maintain a non-degenerate spatial deployment, and (iii) adapt to unknown but bounded variations in the desired heading over time. In our prior work, we provided these guarantees for 2D robots modeled as single integrators and unicycles with constant speeds. 

In this paper, we extend these results to 3D motion by developing a \emph{geometric low-level control framework} on the Lie group $\text{SO}(3)$ for source seeking tasks. Here, we focus on first-order attitude dynamics, working directly on $\text{SO}(3)$ to avoid Euler-angle singularities and quaternion ambiguities, yielding a unique and intrinsic representation of orientation for attitude control \cite{Bullo2005, Chaturvedi2011}. This enables reliable swarm-level coordination, preserves the resilience properties previously demonstrated in 2D, and allows source-seeking robots to operate with full 3D attitude dynamics.

The paper is organized as follows. Section \ref{sec: preliminaries} introduces notation, $\text{SO}(3)$ theory, source-seeking preliminaries, and the problem statement. In Section \ref{sec: geom_control}, we develop a proportional feed-forward controller on $\text{SO}(3)$, prove exponential alignment under unknown attitude variations, and bound inter-agent distances to prevent degenerate formations. Section \ref{sec: simulations} presents numerical simulations. Finally, Section \ref{sec: conclusion} concludes the paper and outlines future research directions.

\section{Preliminaries and Problem Statement}
\label{sec: preliminaries}

For a vector $u \in \mathbb{R}^3$, $\|u\|$ denotes its Euclidean norm. For a matrix $A \in \mathbb{R}^{m \times n}$, $\|A\|_2$ denotes its induced $2$-norm (largest singular value). The operator $\times$ denotes the vector cross product. $I$ is the $3 \times 3$ identity matrix. The unit $2$-sphere is defined as $\mathbb{S}^2 := \{u \in \mathbb{R}^{3}\, | \, \|u\| = 1\}$.

\subsection{The Special Orthogonal group $\text{SO}(3)$} 

The \emph{special orthogonal} group $\text{SO}(3)$ is defined as
\begin{equation}
\label{eq: SO3}
    \text{SO}(3) := \{R \in \mathbb{R}^{3 \times 3} \,|\, R^\top R = I,\, \det(R) = 1\},
\end{equation}
and represents all rotation in 3D. Each $R \in \text{SO}(3)$ is a rotation matrix that preserves distances and orientation, and can be associated with a point in a smooth manifold. As $\text{SO}(3)$ is both a group and a smooth manifold, it forms a \emph{Lie group}. The associated \emph{Lie algebra} is the vector space of skew-symmetric matrices
\begin{equation} \label{eq: SO3_algebra}
    \mathfrak{so}(3) := \{ S \in \mathbb{R}^{3 \times 3} \,|\, S = - S^\top\},
\end{equation}
equipped with the \emph{Lie bracket} $[\cdot,\cdot] : \mathfrak{so}(3) \rightarrow \mathfrak{so}(3)$ given by the matrix commutator
\begin{equation} \label{eq: so3_liebracket}
    [S_1, S_2] = S_1S_2 - S_2S_1.
\end{equation}
The vector space $\mathfrak{so}(3)$ corresponds to the tangent space at the identity of $\text{SO}(3)$. More generally, the tangent space at $R \in \mathrm{SO}(3)$ is expressed as
\begin{equation}
    \mathrm{T}_R\text{SO}(3) := \{R S \, | \, S \in \mathfrak{so}(3)\}.
\end{equation}

In practice, it is often convenient to identify elements of the Lie algebra $\mathfrak{so}(3)$ with vectors in $\mathbb{R}^3$. Euler's rotation theorem guarantees that any rotation $R \in \mathrm{SO}(3)$ can be represented by a \emph{rotation vector} $\tau = [\tau_x, \tau_y, \tau_z]^\top \in \mathbb{R}^3$, whose direction specifies the rotation axis and whose magnitude $\theta = \|\tau\|$ gives the rotation angle. To transform a rotation vector into a skew-symmetric matrix, we use the \emph{hat} map $\wedge : \mathbb{R}^3 \rightarrow \mathfrak{so}(3)$, defined as
\begin{equation} \label{eq: omega_skew}
    \tau^\wedge = \begin{bmatrix} 0 & -\tau_{z} & \tau_{y} \\ \tau_{z} & 0 & -\tau_{x} \\ -\tau_{y} & \tau_{x} & 0 \end{bmatrix} \in \mathfrak{so}(3),
\end{equation}
with inverse \emph{vee} map $\vee : \mathfrak{so}(3) \rightarrow \mathbb{R}^3$, satisfying $(\tau^\wedge)^\vee = \tau$. Notably, $\tau^\wedge u = \tau \times u$, for any $u\in\mathbb{R}^3$.

Once choosen a convenient basis for $\mathfrak{so}(3)$, the \emph{exponential} map $\exp : \mathfrak{so}(3) \rightarrow \text{SO}(3)$ connects the Lie algebra to the Lie group via Rodrigues' formula, i.e.,
\begin{equation} \label{eq: exp}
    \exp(\tau^\wedge) = I + \frac{\sin\theta}{\theta} \tau^\wedge +\frac{1 - \cos\theta}{\theta^2} ({\tau^\wedge})^2.
\end{equation}
Its inverse, the \emph{logarithmic} map $\log : \text{SO}(3) \rightarrow \mathfrak{so}(3)$, recovers the rotation axis and angle as
\begin{equation} \label{eq: log}
    \log(R) =
    \begin{cases}
       \begin{array}{ll}
           0_{3 \times 3} & R = I\\
           \left( \frac{\theta}{2 \sin\theta}  \right) (R - R^\top ) & R \neq I\\
       \end{array},
   \end{cases}
\end{equation}
where $\theta = \arccos((\mathrm{tr}(R)-1)/2)$, so that $\|\log(R)^\vee\| = \theta$. Note that $\log(R)$ is uniquely defined for $\theta \in [0,\pi)$, and becomes singular at $\theta = \pi$, where $\mathrm{tr}(R) = -1$.

\subsection{Geodesic, Logarithmic, and Frobenius Metrics on $\text{SO}(3)$}

The natural metric on $\text{SO}(3)$ is the \emph{geodesic distance}, directly given by the Euclidean norm of the rotation vector $\log(R_1^\top R_2)^\vee$ connecting two elements $R_1,R_2 \in \text{SO}(3)$, i.e.,
\begin{equation} \label{eq: so3_metric_geod}
    d_{\text{SO}(3)}(R_1, R_2) = \|\log(R_1^\top R_2)^\vee\|.
\end{equation}

An alternative metric, which is particularly useful for convergence analysis, is the l\emph{ogarithmic metric}
\begin{equation} \label{eq: so3_metric_log}
    d_{\text{log}}(R_1, R_2) = \|\log(R_1^\top R_2)\|_F,
\end{equation}
where $\|\cdot\|_F$ denotes the \emph{Frobenius norm}. These two metrics are related through
\begin{equation} \label{eq: frob_product}
    \|\tau^\wedge\|_F = \sqrt{\mathrm{tr}({\tau^\wedge}^\top \tau^\wedge)} = \sqrt{2}\|\tau\|,
\end{equation}
which shows that the Frobenius norm on $\mathfrak{so}(3)$ is proportional to the Euclidean norm in $\mathbb{R}^3$.

While both metrics are affected by the singularity of the logarithmic map when $\mathrm{tr}(R_1^\top R_2) = -1$, the following result shows that such singularities can be avoided under typical control conditions.
\begin{lemma} \label{lem: tr_log}
    Let $R_1(t), R_2(t) \in \text{SO}(3)$ be continuous trajectories satisfying
    \begin{equation*}
        d_{\text{SO}(3)}(R_1(t), R_2(t)) \leq d_{\text{SO}(3)}(R_1(0), R_2(0)), \quad \forall t \geq 0.
    \end{equation*}
    If $\mathrm{tr}(R_1(0)^\top R_2(0)) \neq -1$, then $\mathrm{tr}(R_1(t)^\top R_2(t))\neq -1$ for all $t \geq 0$.
\end{lemma}
\begin{proof}
    Since $d_{\text{SO}(3)}(I,R) = \arccos((\mathrm{tr}(R)-1)/2)$ achieves its maximum value $\pi$ only when $\mathrm{tr}(R) = -1$, and the distance between $R_1(t)$ and $R_2(t)$ remains bounded by its initial value, the maximum distance cannot be reached if initially avoided.
\end{proof}

Another interesting metric on $\text{SO}(3)$ is the \emph{Frobenius distance} 
\begin{equation} \label{eq: metric_frob}
    d_{F}(R_1, R_2) = \|R_1 - R_2\|_F.
\end{equation}
This metric satisfies the identity
\begin{equation*}
    \|R_1 - R_2\|_F = 2 \sqrt{2} \left|\sin\left(\frac{\theta}{2}\right)\right|,
\end{equation*}
where $\theta = d_{\text{SO}(3)}(R_1, R_2)$. Using the inequality $\sin(\varphi) \leq \varphi$ for $\varphi \in [0,\pi]$ and \eqref{eq: frob_product}, it follows that
\begin{equation} \label{eq: metric_ident}
    d_{F}(R_1, R_2) \leq \sqrt{2} d_{\text{SO}(3)}(R_1, R_2) = d_{\text{log}}(R_1, R_2),
\end{equation}
establishing a direct connection between the three metrics introduced in this section.

\subsection{Derivatives on $\text{SO}(3)$ and Robot Kinematics} 

\begin{figure}
    \centering
    \includegraphics[trim={1.95cm 1.5cm 1.95cm 7cm}, clip, width=0.8\columnwidth]{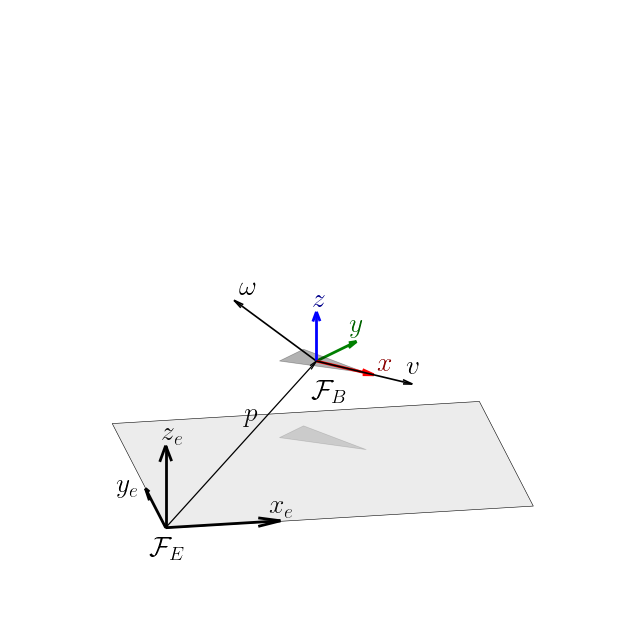}
    \caption{Illustration of a 3D unicycle robot. The robot moves with linear velocity $v =  \begin{bmatrix}s & 0 & 0\end{bmatrix}^\top$ and rotates with an angular velocity $\omega$. These velocities are expressed in the body-fixed frame $\mathcal{F}_B$, which is fixed at the body of the robot, but observed from the inertial frame $\mathcal{F}_E$. Vectors $\{e_x, e_y, e_z\}$ and $\{x_B, y_B, z_B\}$ denote the bases of $\mathcal{F}_E$ and $\mathcal{F}_B$, respectively.}
    \label{fig: fe_vs_fb}
\end{figure}

A robot in 3D space is represented using a \emph{Body-fixed} Cartesian frame $\mathcal{F}_B$. Its position and orientation with respect to an \emph{Earth-fixed} frame $\mathcal{F}_E$ are described by the vector $p \in \mathbb{R}^3$ and the rotation matrix
\begin{equation} \label{eq: R}
   R = \begin{bmatrix}x_B & y_B & z_B\end{bmatrix},
\end{equation}
where $x_B, y_B, z_B \in \mathbb{S}^2$ are the orthonormal unit vectors expressed in $\mathcal{F}_E$ forming the basis of $\mathcal{F}_B$ (see Figure \ref{fig: fe_vs_fb}).

Let $\omega \in \mathbb{R}^3$ denote the robot's angular velocity in the body frame $\mathcal{F}_B$. Differentiating the orthogonality condition $R^\top R = I$ yields $R^\top \dot R$ skew-symmetric, so the rotation matrix kinematics can be expressed as $\dot R = R S \in \mathrm{T}_R\text{SO}(3)$, where $S \in \mathfrak{so}(3)$ encodes the instantaneous rotation in the body frame. Using the same hat map and basis introduced for rotation vectors, we have that $\dot R = R \omega^\wedge$. This representation has a natural connection with the standard angular velocity transformation between frames: the angular velocity expressed in $\mathcal{F}_E$ is $\omega^E = R \omega$, and its skew-symmetric matrix form is $(\omega^E)^\wedge = R \omega^\wedge \in \mathrm{T}_R\text{SO}(3)$.

For a 3D unicycle robot with forward motion constrained along its body $x$-axis, the kinematics are
\begin{equation}
\label{eq: kin}
  \begin{cases}
       \begin{array}{cl}
           \dot p(t) &= R(t) \begin{bmatrix}s & 0 & 0\end{bmatrix}^\top = s x_B \\
           \dot R(t) &= R(t) \Omega \\
       \end{array},
   \end{cases}
\end{equation}
where $\Omega \equiv \omega^\wedge$ is the control input associated with the body angular velocity vector $\omega$. The scalar $s > 0$ denotes the forward speed, which is assumed constant so that the focus remains on attitude control. From this point onward, we adopt $\Omega$ in place of $\omega^\wedge$ to streamline notation.

The \emph{adjoint map} $\mathrm{Ad}_R : \mathfrak{so}(3) \rightarrow \mathfrak{so}(3)$ describes how angular velocities transform under rotations 
\begin{equation} \label{eq: adjoint_map}
    \mathrm{Ad}_R(\Omega) = R \Omega R^\top,
\end{equation}
and satisfies
\begin{equation} \label{eq: adjoint_vee}
    (\mathrm{Ad}_R(\Omega))^\vee = R \Omega^\vee = R\omega.
\end{equation}
That is, the adjoint action on $\mathfrak{so}(3)$ corresponds to the standard rotation of angular velocity vectors in $\mathbb{R}^3$. In this sense, the matrix $R$ can be identified as the \emph{adjoint matrix} $[\mathrm{Ad}_R] : \mathbb{R}^3 \rightarrow \mathbb{R}^3$. This identification is particularly simple in $\text{SO}(3)$ due to the isomorphism between $\mathfrak{so}(3)$ and $\mathbb{R}^3$.

The derivative of the adjoint map is given by the \emph{adjoint operator} $\mathrm{ad}_\Omega : \mathfrak{so}(3) \rightarrow \mathfrak{so}(3)$, which captures the Lie bracket structure of the Lie Algebra, i.e.,
\begin{equation} \label{eq: adjoint_operator}
    \mathrm{ad}_{\Omega_1}(\Omega_2) = [\Omega_1, \Omega_2] = \Omega_1\Omega_2 - \Omega_2\Omega_1.
\end{equation}
Note that $\mathrm{ad}_{\Omega}(\Omega) = 0$ and $[\omega^\wedge_1, \omega^\wedge_2] = (\omega_1 \times \omega_2)^\wedge$. 

The dynamics of the exponential coordinates $\tau^\wedge = \log(R)$ are related to the body angular velocity $\Omega$ through
\begin{align} \label{eq: tau_dot}
    \dot\tau^\wedge = \mathcal{B}_{\tau^\wedge} (\Omega) = \Omega + \frac{1}{2}\mathrm{ad}_{\tau^\wedge}(\Omega) + \frac{1 - \alpha(\theta)}{\theta^2}\mathrm{ad}^2_{\tau^\wedge} (\Omega),
\end{align}
where $\theta = \|\tau\|$ and $\alpha(\theta) := (\theta/2)\cot(\theta/2)$. For small rotations ($\theta \to 0$), $\mathcal{B}_{\tau^\wedge} \to I$ and $\dot \tau^\wedge \to \Omega$.

Finally, a very useful property of the logarithmic metric in \eqref{eq: so3_metric_log} is that it its \emph{Ad-invariant}. That is, for the inner product $\langle \Omega_1, \Omega_2 \rangle = \mathrm{tr}(\Omega_1^\top \Omega_2)$, we have
\begin{equation*}
    \langle \mathrm{Ad}_R(\Omega_1), \mathrm{Ad}_R(\Omega_2) \rangle = \langle \Omega_1, \Omega_2 \rangle.
\end{equation*}
A direct consequence of this invariance is the skew-symmetry identity
\begin{equation} \label{eq: ad_skew}
    \langle \mathrm{ad}_{\Omega_1}(\Omega_2), \Omega_3 \rangle = -\langle \Omega_2, \mathrm{ad}_{\Omega_1}(\Omega_3) \rangle,
\end{equation}
which plays a central role in our Lyapunov analysis. For further background on $\mathrm{SO}(3)$ and Lie groups in robotics, we refer the reader to \cite{bullo_murray_1995, Hall2015, so3_catalanes}.

\subsection{The Ascending Direction and Problem Formulation}

Given a swarm of $N$ robots with centroid $p_c = \frac{1}{N}\sum_i^N p_i$ and barycentric position $x_i \in \mathbb{R}^3$. We define the \emph{deployment} as $x := [x_1^\top \cdots x_N^\top]^\top \mathbb{R}^{3N}$. A deployment is \emph{not degenerated} if the vectors $\{x_i\}$ span $\mathbb{R}^3$, i.e., the covariance matrix $P(x) = \frac{1}{N}\sum_i x_i x_i^\top$ is full rank.

For a scalar field $\sigma: \mathbb{R}^3 \rightarrow \mathbb{R}_+$ of class $C^2$ with a unique maximum at $p_\sigma$, a vector $u \in \mathbb{R}^3$ is an \emph{ascending direction} at $p$ if $\nabla\sigma(p)^\top u > 0$, which ensures approaching to the source. \cite[Proposition 1]{bautista2025resilient} shows that, provided the deployment is non-degenerate and the scalar field has bounded gradient and curvature, then by suitably adjusting the deployment covariance the vector 
\begin{equation} \label{eq: l_sigma}
    L_\sigma(p_c, x) = \frac{1}{ND^2} \sum^{N}_{i=1}\sigma(p_c + x_i)x_i,
\end{equation}
where $D = \max_i\|x_i\|$, is guaranteed to be an ascending direction at $p_c$. 

The adjustment of the deployment covariance in a distributed manner is nontrivial and beyond the scope of this paper. Accordingly, we make the following formal assumption.

\begin{assum} [Suitable deployment covariance] \label{assumption}
    If the deployment $x$ is non-degenerate, the covariance matrix $P(x)$ satisfies the conditions of \cite[Proposition 1]{bautista2025resilient}, ensuring that $L_\sigma(p_c, x)$ is an ascending direction at $p_c$.
\end{assum}

 
Since the robots described by \eqref{eq: kin} move forward along their body $x$-axis $x_B$ at a constant speed, the magnitude of $L_\sigma$ is not relevant for source seeking in this work. We therefore define 
\begin{equation} \label{eq: md}
    m_d(p) = \frac{L_\sigma(p)}{\|L_\sigma(p)\|} \in \mathbb{S}^2
\end{equation}
as the vector field to be tracked by each robot in the swarm. Note that $m_d$ is tangent to the unit sphere, i.e., $m_d^\top \dot m_d = 0$, so $\dot m_d$ describes how fast the vector rotates on the unit sphere. The rotation of $m_d$ can be expressed as $\dot m = \omega_m \times m$, where $\omega_m \in \mathbb{R}^3$ is the instantaneous angular velocity vector of $m$. Then, $\|\dot m\| = \|\omega_m\|$, since $\|m\|=1$ and $\omega_m$ is perpendicular to $m$.

The alignment error between a robot's heading and the ascending direction is quantified using the geodesic distance on the sphere $\mathbb{S}^2$, defined as
\begin{equation} \label{eq: dis_s2}
    d_{\mathbb{S}^2}(u_1, u_2) = \arccos(u_1^\top u_2) \in [0,\pi],
\end{equation}
where $u_1,u_2 \in \mathbb{S}^2$

Finally, note that, under Assumption \ref{assumption}, $m_d(p)$ is an ascending direction at $p$ only if the deployment $x$ is non-degenerate. Thus, the source-seeking task in this work requires both alignment with $m_d$ and preservation of a non-degenerate deployment, and can therefore be reduced to the following problem.

\begin{problem} \label{prolem: main}
Consider a swarm of $N$ 3D unicycle robots with dynamics given by \eqref{eq: kin}. Design the actuation $\Omega(t)$ in \eqref{eq: kin}, via the control input $\omega(t)$, such that if the initial deployment $x(0)$ is non-degenerate, and $\|\dot m_d\| \leq \omega_m^\text{max}$ with $\omega_m^\text{max} \geq 0$, the following conditions hold:
\begin{enumerate}[label=(\roman*)]
    \item Each robot $i$ aligns its heading with the desired direction, i.e., there exist a $\delta^*>0$ such that $d_{\mathbb{S}^2}(x_{B_i}(t), m_d(p(t))) \rightarrow [0,\delta^*]$ exponentially fast as $t \to \infty$.
    \item The deployment $x(t)$ remains non-degenerate $\forall t \geq 0$.
\end{enumerate}
\end{problem}

\begin{remark}
    Note that aligning every robot with $m_d$ is sufficient, though not strictly necessary, for source-seeking, as the robots could instead coordinate their motions to keep the centroid moving along $m_d$. Nevertheless, the simpler approach is preferred in practice, as it reduces the complexity of the low-level controller while also helping to preserve inter-vehicle distances and the deployment covariance. 
\end{remark}




\section{$\text{SO}(3)$ Geometric Control for Ascending Direction Tracking}
\label{sec: geom_control}

\subsection{Alignment with a Generic Desired Attitude}

We define the \emph{desired attitude} as the rotation matrix
\begin{equation} \label{eq: Rd}
   R_d := \begin{bmatrix} x_{B_d} & y_{B_d} & z_{B_d} \end{bmatrix} \in \text{SO}(3),
\end{equation}
where $x_{B_d}, y_{B_d}, z_{B_d} \in \mathbb{S}^2$ are the orthonormal unit vectors expressed in the earth-fixed frame $\mathcal{F}_E$, collectively describing the desired orientation. In general, $R_d$ may vary with both position and time, i.e., $R_d = R_d(p,t)$; however, this dependence will be omitted whenever clear from context. Due to the orthogonality constraint, specifying any one of the vectors in $R_d$ uniquely restricts the other two to lie on the orthogonal plane.

In this subsection, we consider a generic $R_d$, assuming that its time variation $\dot R_d$ is known. The specific case where $x_{B_d}$ corresponds to the ascending direction $L_\sigma$, with $y_{B_d}$ and $z_{B_d}$ chosen by design to complete a right-handed basis, will be addressed in the following subsection.

Using the geodesic distance in \eqref{eq: so3_metric_geod}, the \emph{attitude error} signal is given by
\begin{equation} \label{eq: mu_r}
   \mu_{R_e} := d_{\text{SO}(3)}(R_d, R) = \|\log(R_e)^\vee\|,
\end{equation}
where $R_e = R_d^\top  R \in \text{SO}(3)$ denotes the \emph{attitude error matrix}. Using this same metric, the notion of exponential stability can be naturally extended to trajectories on $\text{SO}(3)$ as follows.

\begin{defn}
\label{def: exptt} [Local Exponential Stability in Trajectory Tracking on $\text{SO}(3)$]
    Consider a trajectory $R(t) \in \text{SO}(3)$ and a desired reference trajectory $R_d(t) \in \text{SO}(3)$. The tracking of $R_d(t)$ is \emph{exponentially stable} if there exist constants $\epsilon, \alpha, k > 0$ such that for all initial conditions satisfying $d_{\text{SO}(3)}(R(t_0), R_d(t_0)) \leq \epsilon$, the bound 
    \begin{equation} \label{eq: exp_stability}
        d_{\text{SO}(3)}(R(t), R_d(t)) \leq \alpha d_{\text{SO}(3)}(R(t_0), R_d(t_0)) e^{-k(t-t_0)}
    \end{equation}
    holds for all $t \geq t_0$.
\end{defn}

Introducing $R_e$ in the control input of \eqref{eq: kin}, i.e., letting $\Omega = \Omega(R_e)$, yields the feedback system 
\begin{equation} \label{eq: kin_fb}
    \begin{cases}
        \begin{array}{cl}
            \dot p &= R \begin{bmatrix}s & 0 & 0\end{bmatrix}^\top \\
            \dot R &= R \Omega(R_e) \\
            \dot R_e &= R_e \Omega_e
        \end{array},
    \end{cases}
\end{equation}
where $\Omega_e \in \mathfrak{so}(3)$ is the angular velocity tensor associated with the time variation of the attitude error matrix $R_e$. This term depends on the time variation of the desired attitude, given by $\Omega_d \in \mathfrak{so}(3)$.


Differentiating $R_e = R_d^\top R$ gives
\begin{align} \label{eq: R_e_dot}
    \dot R_e &= R_d^{\top}\dot R + \frac{\mathrm{d}}{\mathrm{dt}}(R_d^{\top})R \nonumber\\  
    &= R_d^{\top} \dot R - R_d^{\top} \dot R_d R_d^{\top} R \nonumber\\
    &= R_e(\Omega - R^{\top}\dot R_d R_d^{\top} R),
\end{align}
where we have used the identity $\frac{\mathrm{d}}{\mathrm{dt}}(R^{\top}) = R^{\top}\dot R R^{\top}$. Comparing \eqref{eq: R_e_dot} with $\dot R_e = R_e \Omega_e$ yields the angular velocity tensor of the error dynamics as
\begin{align} \label{eq: omega_e}
    \Omega_e &= \Omega - R^{\top}\dot R_d R_d^{\top} R \nonumber\\
    &= \Omega - R_e^{\top} \Omega_d R_e \nonumber\\
    &= \Omega - \mathrm{Ad}_{R_e^{\top}}(\Omega_d),
\end{align}
and by applying the vee map (while using \eqref{eq: adjoint_vee}) we obtain the equivalent vector form
\begin{align} \label{eq: omega_e_vee}
     \Omega_e^\vee = \Omega^\vee - R_e^{\top}\Omega_d^\vee.
\end{align}

Equation \eqref{eq: omega_e_vee} shows that the error dynamics can be expressed directly as a vector in $\mathbb{R}^3$, representing the difference between the robot's angular velocity and the desired attitude time variation, mapped into the body frame through the adjoint transformation $R_e^{\top}$. Thus, operating with $\Omega_e$ in $\mathfrak{so}(3)$ or with $\Omega_e^\vee$ in $\mathbb{R}^3$ conveys the same information

Once the error signal $\mu_{R_e}(t)$ and the feedback system \eqref{eq: R_e_dot} have been introduced, we are ready to formalize a control law that ensures a robot governed by \eqref{eq: kin} converges to a time-varying desired attitude $R_d(t)$.

\begin{prop}[3D Attitude Controller] \label{prop: att_control}
    Consider the desired attitude trajectory $R_d(t)$ and the feedback system \eqref{eq: kin_fb}. For any initial condition $R(t_0)$ such that $\mathrm{tr}(R_e(t_0)) \neq -1 $, the tracking of $R_d(t)$ is exponentially stable under the control law
    \begin{align} \label{eq: omega_control}
        \Omega(R_e) &= - k_w \log(R_d^{\top}R) + R^{\top} \dot R_d R_d^{\top} R \nonumber \\
        &= - k_w \log(R_e) + \mathrm{Ad}_{R_e^\top } (\Omega_d),
    \end{align}
    where $k_w > 0$ is a proportional gain modulating the convergence rate.
\end{prop}
\begin{proof}
    Consider the Lyapunov candidate $V(R_e) = \mu_{R_e}^2$, which considering \eqref{eq: frob_product} yields
    \begin{equation} \label{eq: V}
        V(R_e) = \frac{1}{2} \|\log(R_e)\|_F^2 = \frac{1}{2} \mathrm{tr}(\log(R_e)^\top \log(R_e)),
    \end{equation}
    which is well-defined whenever $\mathrm{tr}(R_e) \neq -1$. Differentiating with respect to time gives
    \begin{align} \label{eq: V_dot_B}
        \dot V(R_e) &= \mathrm{tr}(\log(R_e)^\top \frac{\mathrm{d}}{\mathrm{dt}}\log(R_e)) \nonumber\\
        &= \mathrm{tr}(\log(R_e)^\top \mathcal{B}_{\log(R_e)}(\Omega_e)) \nonumber \\
        &= \mathrm{tr}(\log(R_e)^\top  \Omega_e),
    \end{align}
    where we have used \eqref{eq: tau_dot} and \eqref{eq: ad_skew}. In particular, the identity $\mathrm{ad}_{\log(R_e)}(\log(R_e)) = 0$ imply that the following inner products vanish:
    \begin{align*}
        \langle \mathrm{ad}_{\log(R_e)}(\Omega) , \log(R_e) \rangle &= - \langle \Omega , \mathrm{ad}_{\log(R_e)}(\log(R_e)) \rangle = 0,\\
        \langle \mathrm{ad}^2_{\log(R_e)}(\Omega) , \log(R_e) \rangle &= \langle \Omega , \mathrm{ad}^2_{\log(R_e)}(\log(R_e)) \rangle = 0,
    \end{align*}
    where we recall that $\langle \Omega_1, \Omega_2 \rangle = \mathrm{tr}(\Omega_1^\top \Omega_2)$.

    Therefore, choosing
    \begin{align} \label{eq: omega_e_kw}
        \Omega_e = - k_w \log(R_e),
    \end{align}
    yields $\dot V(R_e) = - 2k_w V(R_e)$. The solution of this ODE is $V(R_e(t)) = V(R_e(t_0)) e^{-2k_w t}$, which implies 
    $$\mu_{R_e}(t) = \mu_{R_e}(t_0) e^{-k_w t}.$$
    Thus, by Definition \ref{def: exptt}, and Lemma \ref{lem: tr_log}, the tracking of $R_d(t)$ is exponentially stable whenever $\mathrm{tr}(R_e(t_0)) \neq -1$. Finally, substituting \eqref{eq: omega_e_kw} into \eqref{eq: omega_e} recovers the control law \eqref{eq: omega_control}.
\end{proof}

The control law \eqref{eq: omega_control} consists of a proportional term that reduces the attitude error and a feed-forward term compensating for the time variation of the desired attitude. Note that if $R_d$ is constant, then $\Omega_d = 0$ and the control law reduces to a pure proportional controller. However, when $\Omega_d$ is unknown, this results in a disturbance term that can prevent convergence. This is the case when aligning to the ascending direction $L_\sigma$ of an unknown scalar field $\sigma(p)$. In the next subsection, we will address this challenge by first bounding $\Omega_d$ and then designing a robust controller to handle the uncertainty.

\subsection{Feedback Compensation for Unknown but Bounded Desired Attitude}

We now consider the case where the desired heading $m_d(t)$, defined in \eqref{eq: md}, evolves over time with an unknown but bounded rate of change, i.e., $\|\dot m_d(t)\| \leq \omega_m^\text{max}$. The control objective is to align the heading direction of a robot, described by \eqref{eq: kin}, with $m_d(t)$, which is equivalent to minimize the geodesic distance \eqref{eq: dis_s2} between $x_B(t)$ and $m_d(t)$ on the unit sphere. Therefore, rather than converging to a specific desired attitude trajectory $R_d(t)$, the objective of this section is to converge to a \emph{desired attitude set}, defined as follows.

\begin{defn} [Desired Attitude Set]
    Let $R \in \text{SO}(3)$ be as defined in \eqref{eq: R}, and let $R_d(t) \in \text{SO}(3)$ in \eqref{eq: Rd} be such that $x_{B_d}(t) = m_d(t)$, with $y_{B_d}, z_{B_d} \in \mathbb{S}^2$ chosen to complete a right-handed orthonormal basis.
    Given a constant $\delta \in [0,\pi]$, the desired attitude set at time $t$ is defined as
    \begin{equation*}
        \mathcal{R}_\delta(t) := \{R \in \text{SO}(3) \,|\, d_{\mathbb{S}^2}(x_B, m_d(t)) \leq \delta\}.
    \end{equation*}
\end{defn}

\begin{remark}
    For the kinematics in \eqref{eq: kin}, only the alignment of the body $x$-axis is relevant for source seeking, since the robot moves forward along $x_B$. The choice of $y_{B_d}$ and $z_{B_d}$ is therefore arbitrary in this context. However, in more realistic scenarios, these axes could be used to encode secondary objectives, e.g., maintaining a prescribed roll or pitch angle for aerodynamic stability.
\end{remark}

Firs, note that for $R_d(t)$ as in \eqref{eq: Rd}, with $x_{B_d}(t) = m_d(t)$, the time variation of $R_d(t)$ can be expressed as
\begin{equation} \label{eq: omega_a_unk}
    \Omega_d = \Omega_d^u + \Omega_d^k,
\end{equation}
where $\Omega_d^u \in \mathfrak{so}(3)$ represents the \emph{unknown component} from $m_d(t)$ variation, bounded as $\|{\Omega_d^u}^\vee\| \leq \omega_m^\text{max}$, and $\Omega_d^k \in \mathfrak{so}(3)$ denotes the \emph{known component} determined by the choice of the complementary axes $y_{B_d}(t)$ and $z_{B_d}(t)$, which may be time-varying but are specified by design. 

Since $\Omega_d^u$ cannot be measured, we reformulate the control law in \eqref{eq: omega_control} to depend solely on the known component, i.e.,
\begin{equation} \label{eq: omega_control_know}
    \Omega(R_e) = - k_\omega \log(R_e) + \mathrm{Ad}_{R_e^\top} (\Omega_d^k),
\end{equation}
which will serve as the foundation for the following technical result.

\begin{prop} \label{prop: exp_stab}
    Consider the feedback system \eqref{eq: kin_fb} with a desired attitude trajectory $R_d(t)$ whose time variation $\Omega_d$ satisfies the decomposition \eqref{eq: omega_a_unk}. Assume that the unknown component obeys $\|{\Omega_d^u}^\vee\| < \omega_m^\text{max}$ for some constant $\omega_m^\text{max} \geq 0$. If $\mathrm{tr}(R_e(t_0)) \neq -1$, then the control law \eqref{eq: omega_control_know} ensures exponential convergence of $R(t)$ to the set $\mathcal{R}_{\delta^*}(t)$, provided the proportional gain $k_\omega$ is selected as
    $$k_\omega = \sqrt{2}\frac{\omega_m^\text{max}}{\mu^*_{R_e}},$$
    with $\mu^*_{R_e} \leq \delta^*$.
\end{prop}

\begin{proof} Substituting the decomposition \eqref{eq: omega_a_unk} and the control law \eqref{eq: omega_control_know} into the error dynamics \eqref{eq: omega_e_vee} yields
\begin{align} \label{eq: omega_e_unk}
    \Omega_e^\vee = - k_w \log(R_e)^\vee - {R_e^\top}^\vee ({\Omega_d^u})^\vee.
\end{align}
Consider the Lyapunov function $V(R_e) = \frac{1}{4}\mu_{R_e}^2$. Following the derivation in \eqref{eq: V_dot_B} and using \eqref{eq: frob_product}, its time derivative along the system trajectories can be expressed as
\begin{align*}
    \dot V (R_e) = \frac{1}{2}\mathrm{tr}\left(\log(R_e)^\top \Omega_e\right) = \tau_e^\top \Omega_e^\vee,
\end{align*}
where $\tau_e = \log(R_e)^\vee$. Substituting $\Omega_e^\vee$ from \eqref{eq: omega_e_unk} gives
\begin{align} \label{eq: to_max}
    \dot V (R_e) = - \frac{1}{2}k_w \mu_{R_e}^2 - \tau_e^\top \left(R_e^\top (\Omega_d^u)^\vee\right),
\end{align}
This separates the stabilizing term from the disturbance induced by $\Omega_d^u$. Using Cauchy-Schwarz and the fact that $\|\tau_e\| = \frac{1}{\sqrt{2}} \mu_{R_e}$, the disturbance introduced by the second term in \eqref{eq: to_max} can be bounded as
$$\tau_e^\top \left(R_e^\top (\Omega_d^u)^\vee\right) \leq \frac{\mu_{R_e}}{\sqrt{2}}\omega_m^\text{max}.$$
Hence, defining $C(R_e) := \left(-k_\omega + \sqrt{2}\frac{\omega_m^\text{max}}{\mu_{R_e}} \right)$, we can write
\begin{align*}
    \dot V (R_e) \leq \frac{1}{2} C(R_e) \mu_{R_e}^2 = 2 C(R_e) V(R_e).
\end{align*}
By choosing the gain as $k_\omega = \sqrt{2}\omega_m^\text{max}/\mu^*_{R_e}$, we ensure that $C(R_e) < 0$ whenever $\mu_{R_e} > \mu^*_{R_e}$. Therefore, the Lyapunov function $V(R_e)$ decreases exponentially until $\mu_{R_e} \leq \mu_{R_e}^*$. As a result, $\mu_{R_e}(t)$ converges exponentially fast to the interval $[0, \mu_{R_e}^*]$, guaranteeing that the attitude error remains bounded by $\mu_{R_e}^*$. Setting $\mu_{R_e}^* \leq \delta^*$, it is equivalent to say that $R(t)$ converges exponentially to the set $\mathcal{R}_{\delta^*}(t)$. To see why this choice is sufficient, consider the following argument.

First, recall that $\tau_e = \log(R_e)^\vee$ is the rotation vector connecting $R$ to $R_d$, which is equal to the rotation angle $\theta_e = \|\tau_e\| = \mu_{R_e}$ about a unit axis $l_e \in \mathbb{R}^3$. Aplying this rotation to the body $x$-axis $x_B$ through the Rodrigues' rotation formula gives
\begin{align*}
    x_B^\text{rot} = x_{B}^\parallel + \cos(\theta_e) x_B^\perp + \sin(\theta_e)l_e \times x_B^\perp,
\end{align*}
where $x_B^\parallel = (x_B^\top l_e)l_e$ is the component of $x_B$ parallel to $l_e$, and $x_B^\perp = x_B - x_B^\parallel$ is the component orthogonal to $l_e$. Hence, since $x_B^\text{rot} = x_{B_d} = m_d$, we have
\begin{align*}
    \cos(\delta) = x_B^\top m_d &= \|x_B^\parallel\|^2 + \cos(\theta_e)\|x_B^\perp\|^2\\
    &= \cos(\theta_e) + \|x_B^\parallel\|^2(1 - \cos(\theta_e)),
\end{align*}
so $\cos(\delta) \geq \cos(\theta_e) = \cos(\mu_{R_e})$ and consequently $\delta \leq \mu_{R_e}$. Hence, $\mu^*_{R_e} \leq \delta^*$ ensures that $\delta \leq \delta^*$.
\end{proof}

The controller in \eqref{eq: omega_control_know}, with $k_w$ chosen according to Proposition \ref{prop: exp_stab}, allows compensation for bounded variations in the desired attitude by increasing the gain $k_w$. This addresses the first objective of Problem \ref{prolem: main}. 

\subsection{Deployment Dispersion Analysis During Transients}

We will finish this section by showing how to ensure that the deployment $x$ remains non-degenerate during transients. To this end, we leverage the exponential convergence of the attitude error $\mu_{R_e}(t)$ established in Proposition \ref{prop: exp_stab} to compute the following pairwise displacement bound.

\begin{lemma} \label{lem: pij}
    Consider two robots $i, j$ with dynamics \eqref{eq: kin}, sharing the same desired attitude matrix $R_d \in \text{SO}(3)$. If the control input $\Omega_{\{i,j\}}$ is designed as in \eqref{eq: omega_control}, then for any initial condition $R(t_0)$ satisfying $\mathrm{tr}(R_e(t_0)) \neq -1 $, the relative position between the two robots satisfies
    \begin{equation*}
        \|p_{ij}(t) - p_{ij}(t_0)\| \leq \frac{2\pi s}{k_w},
    \end{equation*}
    for all $t \geq t_0$.
\end{lemma}
\begin{proof}
    From \eqref{eq: metric_ident} and Proposition \ref{prop: att_control}, for each robot $i$ we have
    \begin{equation} \label{eq: exp_decay}
        \|R_i(t) - R_d(t)\|_F \leq \mu_{R_e^i}(t) \leq \mu_{R_e^i}(t_0) e^{-k_w t},
    \end{equation}
    where $R_e^i = R_i^\top R_d$. Using the triangle inequality, the distance between the two robots' attitudes satisfies
    $$
    \|R_i(t) - R_j(t)\|_F \leq (\mu_{R_e^i}(t_0) + \mu_{R_e^j}(t_0)) e^{-k_w t} \leq 2\pi e^{-k_w t}.
    $$
    
    Given $\dot p_{ij} = (R_i - R_j)v$ and $\|v\| = s$, we have
    \begin{equation*}
        \|\dot p_{ij}\| \leq \|v\| \|R_i - R_j\|_F = s \|R_i - R_j\|_F \leq (2\pi s) e^{-k_w t}.
    \end{equation*}
    Finally, integrating $\dot p_{ij}$ over time gives
    \begin{align*}
        \|p_{ij}(t) - p_{ij}(t_0)\| &= \left\|\int_{t_0}^t  \dot p_{ij}(s) \mathrm{ds}\right\| \leq \int_{t_0}^t  \|\dot p_{ij}(s)\| \mathrm{ds}\\ &\leq 2\pi s\int_0^\infty  e^{- k_w s} \mathrm{ds} =  \frac{2\pi s}{k_w},
    \end{align*}
    which completes the proof
\end{proof}

\begin{remark}
Note that Lemma \ref{lem: pij} also holds when the variations of $R_d(t)$ are unknown but common to all robots. Since every agent tracks the same $R_d(t)$, their attitude errors still decay exponentially toward this reference, as assumed in \eqref{eq: exp_decay}. Therefore, even if the variation is too fast to be tracked exactly (e.g., $\|\Omega_d^u\| > \omega_m^{\max}$), all robots converge to the same displaced equilibrium point close to $R_d(t)$, and the formation is preserved.
\end{remark}

Lemma \ref{lem: pij} itself is not sufficient to ensure that the deployment $x$ remains non-degenerate, as it only bounds the pairwise displacements. However, it can be used to derive the following sufficient condition.

\begin{prop} \label{prop: non_deg}
    Suppose the initial deployment $x(t_0)$ is non-degenerate, and assume the pairwise displacement bound in Lemma \ref{lem: pij} holds. Then the deployment $x(t)$ remains non-degenerate for all $t \geq 0$ provided that
    \begin{equation*}
        k_w > \frac{2 \pi s}{-D_0 + \sqrt{D_0^2 + \lambda_{\text{min}}(P(t_0))}},
    \end{equation*}
    where $D_0 := \max_i\|x_i(t_0)\|$ and $\lambda_{\text{min}}(P(t_0))$ denotes the minimum eigenvalue of the initial covariance matrix $P(t_0)$.
\end{prop}
\begin{proof}
    We need a condition on the pairwise displacement bound $\|p_{ij}(t) - p_{ij}(t_0)\| \leq \epsilon$, with $\epsilon = \frac{2\pi s}{k_w}$, that guarantees the centroid-relative covariance $P(t)$ remains full rank, i.e., $\lambda_{\text{min}}(P(t)) > 0$, for all $t \geq t_0$. 
    
    Let express $P(t)$ as a perturbation of $P(t_0)$, i.e., $P(t) = P(t_0) + \Delta P(t)$ with
    \begin{equation*}
        \Delta P(t) := \frac{1}{N} \sum_{i=k}^N (x_k(t)x_k(t)^\top - x_k(t_0)x_k(t_0)^\top).
    \end{equation*}
    Since all matrices are symmetric, using Weyl's inequality \cite[Theorem 4.3.1]{horn2013matrix} gives
    \begin{align*}
        \lambda_{\text{min}}(P(t)) 
        &\geq \lambda_{\text{min}}(P(t_0)) + \lambda_{\text{min}}(\Delta P(t)) \nonumber \\ 
        &\geq \lambda_{\text{min}}(P(t_0)) - \|\Delta P(t)\|_2,
    \end{align*}
    where we have used the fact that $|\lambda_{\text{min}}(\Delta P(t))| \leq \|\Delta P(t)\|_2$ since $\Delta P(t)$ is symmetric. Hence, a sufficient condition to keep $P(t)$ full rank is
    \begin{equation} \label{eq: suff_cond}
        \|\Delta P(t)\|_2 < \lambda_{\text{min}}(P(t_0)).
    \end{equation}
    Next, let us show how to bound $\|\Delta P(t)\|_2$ in terms of the pairwise bound $\epsilon$.

    Firstly, let adopt $\Delta u := \|u(t) - u(t_0)\|$, $u\in\mathbb{R}^3$, to streamline notation. The hypothesis $\|\Delta p_{ij}\| < \epsilon$ for all $i,j$ implies $\|\Delta p_i - \Delta p_j\| < \epsilon$. For the centroid-relative coordinates we have that $\Delta x_i = \Delta p_i - \Delta p_c$, where $\Delta p_c = \frac{1}{N}\sum_{i=k}^N \Delta p_k$. Thus, using the triangle inequality gives
    \begin{align*} \label{eq: delta_xi}
        \|\Delta x_i\| = \|\Delta p_i - \Delta p_c\| 
        & \leq \frac{1}{N} \sum_{k=1}^N \|\Delta p_i - \Delta p_k\| \leq \epsilon.
    \end{align*}
    
    Expanding the definition of $\Delta P(t)$ with $x_k(t) = x_k(t_0) + \Delta x_k$ gives
    \begin{equation*}
        \Delta P(t) = \frac{1}{N} \sum_{i=1}^N \|x_k(t_0)\Delta x_i^\top + \Delta x_i x_i(t_0)^\top + \Delta x_i \Delta x_i^\top\|_2.
    \end{equation*}
    Applying the triangle inequality and using $\|\Delta x_i\| \leq \epsilon$ yields
    \begin{equation*}
        \|\Delta P(t)\|_2 = \frac{1}{N} \sum_{i=1}^N (2 \|x_i(t_0)\|\|\Delta x_i\| + \|\Delta x_i\|^2) \leq 2 D_0 \epsilon + \epsilon^2.
    \end{equation*}
    Therefore, to satisfy the sufficient condition \eqref{eq: suff_cond}, we require $2 D_0 \epsilon + \epsilon^2 \leq \lambda_{\text{min}}(P(t_0))$. Solving this quadratic inequality for $\epsilon > 0$ gives $\epsilon < \epsilon_\text{max} := -D_0 + \sqrt{D_0^2 + \lambda_{\text{min}}(P(t_0))}$. Recalling that $\epsilon = \frac{2\pi s}{k_w}$, this yields the lower bound on the control gain $k_w > \frac{2\pi s}{\epsilon_\text{max}}$, which completes the proof.
\end{proof}

\begin{remark}
    Distributed estimation of the covariance matrix $P(t)$ can be achieved using consensus-based algorithms, as discussed in \cite{chen2025dispersion}. This approach allows each robot to compute and update its local estimate of $P(t)$ based on information exchanged with neighboring robots, ensuring that the entire swarm maintains an accurate representation of the deployment's covariance without relying on a centralized controller.
\end{remark}

The convergence analysis in Proposition \ref{prop: att_control}, along with the proportional gain bounds specified in Proposition \ref{prop: exp_stab} and Proposition \ref{prop: non_deg}, allow us to present the main result of this section, which provides a formal solution to Problem \ref{prolem: main}.

\begin{thm} \label{thm}
    Consider a swarm of $N$ 3D unicycle robots with dynamics \eqref{eq: kin}. Assume the initial deployment $x(t_0)$ is non-degenerate and that the desired attitude trajectory $R_d(t) \in \text{SO}(3)$, is shared by all robots, with $x_{B_d}(t) = m_d(t)$ satisfying $\|\dot m_d(t)\| \leq \omega_m^{\text{max}}$ for some $\omega_m^{\text{max}} \geq 0$. Also, assume that for every robot $i$, the initial relative attiude satisfies $\mathrm{tr}(R_e^i(t_0)) \neq -1$, where $R_e^i = R_i^\top R_d$. Let $k_1,k_2 > 0$ be the proportional gain lower bounds specified in Proposition \ref{prop: exp_stab} and Proposition \ref{prop: non_deg}, respectively. 
    If each robot applies the control law \eqref{eq: omega_control_know} with $k_w = \max\{k_1,k_2\}$, then Problem \ref{prolem: main} is solved.
\end{thm}
\begin{proof}
    The result follows directly from Proposition \ref{prop: exp_stab} and Proposition \ref{prop: non_deg}. By choosing $k_w = \max\{k_1,k_2\}$, both lower bounds are satisfied simultaneously. Consequently, each $x_{B_i}(t) \to \mathcal{R}_{\delta^*}(t)$ exponentially fast while the deployment remains non-degenerate, so Problem \ref{prolem: main} is solved.
\end{proof}


\section{Simulations}
\label{sec: simulations}

\begin{figure}[t]
    \centering
    \includegraphics[trim={0cm 0cm 0cm 0cm}, clip, width=0.98\columnwidth]{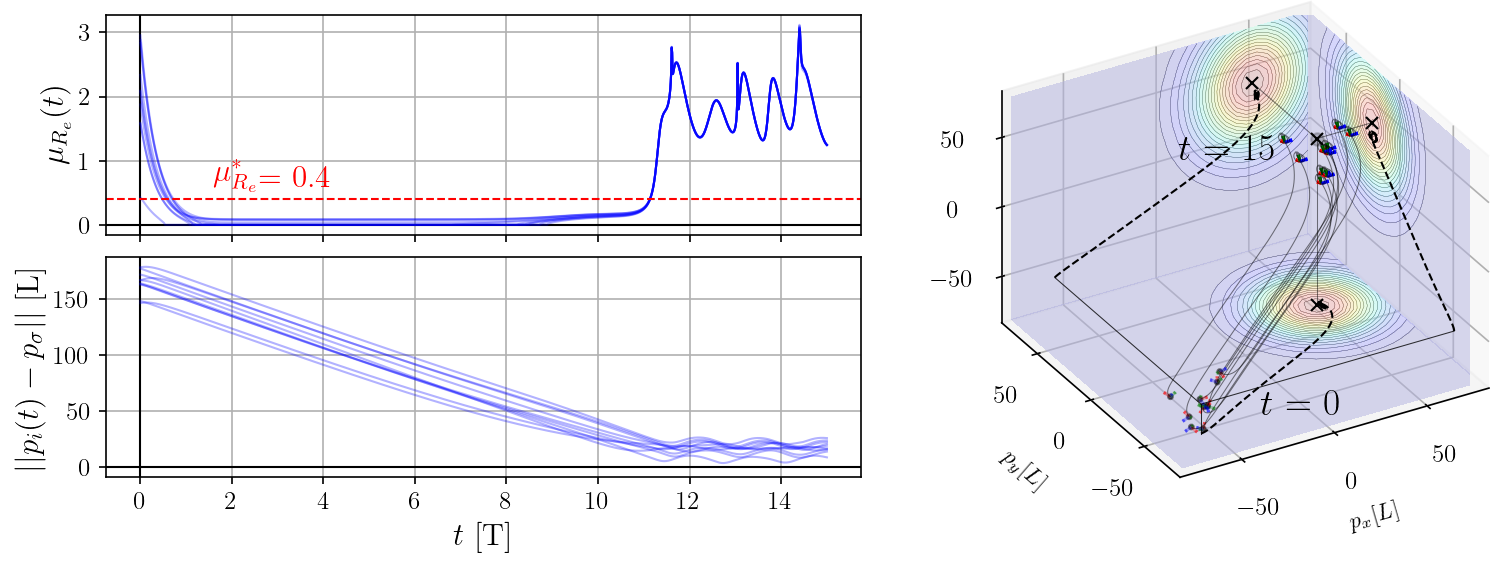}
    \caption{
        A swarm of $N=10$ unicycle robots with constant speed $s = 15$ \textit{space unit/time unit} align their body axis $x_B$ with the desired direction $x_{B_d}(t)=m_d(t)$ from \cite{bautista2025resilient} to seek the source of a scalar field $\sigma(p)$. The desired attitude evolves with angular velocity $w_d(t)=R_d(t)^\top w^k + w^u$, where $\omega^k = [\pi,0,0]$ \textit{radians/time unit} is known and $w^u \in \mathbb{R}^3$ depends on the field and positions, as it is deeply explained in \cite{bautista2025resilient}. 
        Robots assume $\|w^u\| \leq \omega_m^{\text{max}} = \pi/4$ \textit{radians/time unit}, which holds when $p_c(t)$ is in a certain set where $\|\nabla\sigma(p_c(t))\| > \epsilon$ for a given $\epsilon > 0$ \cite[Proposition 8]{bautista2025resilient}. Alignment uses \eqref{eq: omega_control_know} with $k_\omega = \sqrt{2} \omega_m^{\text{max}} / \mu_{R_e}^*$ and $\mu_{R_e}^* = 0.4$ \textit{radians} for all robots. Left: attitude error $\mu_{R_e}(t)$ and distance $\|p_i(t) - p_\sigma\|$ to the source. Right: scalar field projection, robot trajectories, and body axes ($x_{B_i}$, red, $y_{B_i}$, green, $z_{B_i}$, blue).
    }
    \label{fig: sim_ss}
\end{figure}

We validate our results through two simulations. Figure \ref{fig: thm} verifies Theorem \ref{thm} using $k_w = k_1$ instead of $k_w = \max\{k_1,k_2\}$ to demonstrate that the conditions are \emph{sufficient but not necessary}: the alignment error $\delta$ converges exponentially to $[0,\delta^*]$, while the deployment remains non-degenerate, i.e., $\Delta \lambda_\text{min}(t) = \lambda_\text{min}(P(t)) - \lambda_\text{min}(P(0)) \geq - \lambda_\text{min}(P(0))$. With $k_1 = 0.55$ and $k_2 = 413$ (from $\lambda_\text{min}(P(0)) = 0.07$, $D_0 = 3.87$), this confirms the conservativeness of the sufficiency condition—lower gains may suffice in practice. Figure \ref{fig: sim_ss} applies our framework to a 3D source-seeking problem, using the algorithm proposed in \cite{bautista2025resilient} as a high-level control. Source code: \url{github.com/jesusBV20/source_seeking_3D}

\begin{figure}[t]
    \centering
    \includegraphics[trim={0cm 0.2cm 0cm 0cm}, clip, width=0.98\columnwidth]{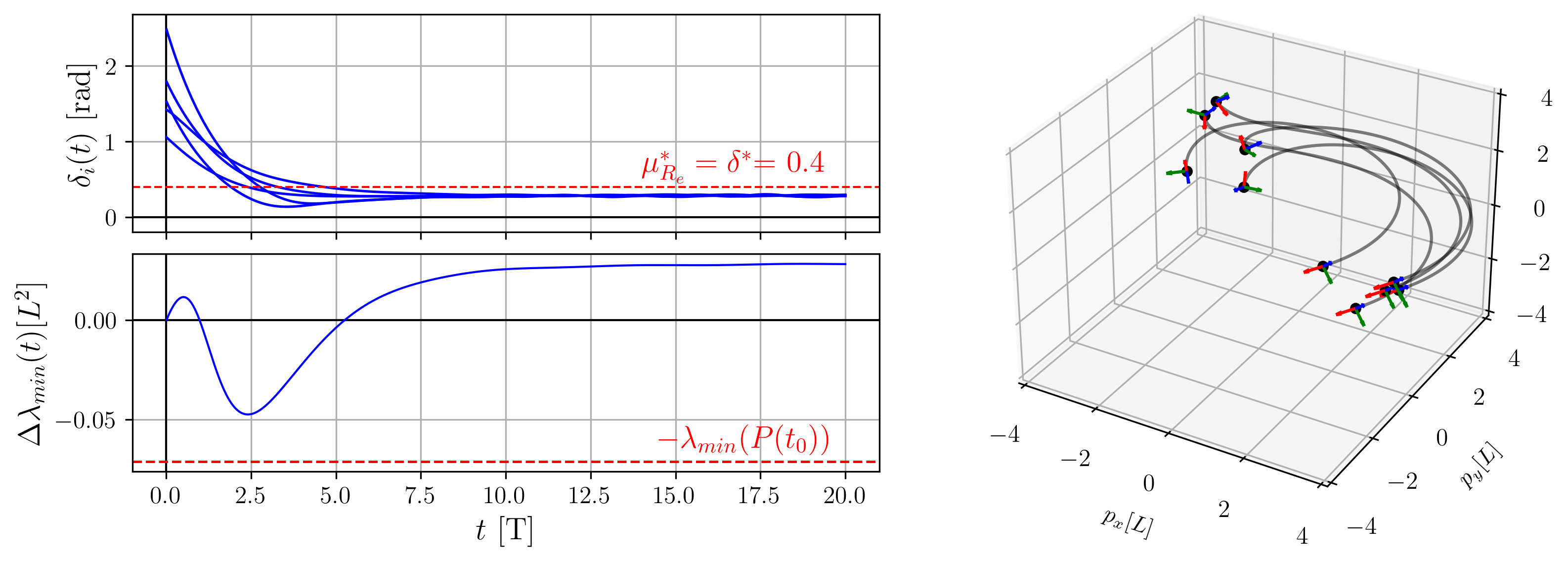}
    \caption{
        A swarm of $N=4$ 3D unicycle robots moving at a constant speed $s=0.6$ \textit{space unit/time unit} along $x_B$, tracking a time-varying desired attitude $R_d(t)$ with $x_{B_d}(t) = m_d(t)$. The time variation of $R_d(t)$ is given by the earth-fixed angular velocity vector $w_d(t) = R_d(t)^\top w^k + w^u$, where $\omega^k = [\pi/2,0,0]$ and $\omega^u = [0,0,-\pi/20]$, both in \textit{radians/time unit}. The known component is $w^k$, while the unknown one satisfies $\|\omega^u\| \leq \omega_m^{\text{max}} = \pi/20$. Alignment control is given by \eqref{eq: omega_control_know} with gain $k_\omega = \sqrt{2}\omega_m^{\text{max}}/\mu_{R_e}^*$ and $\mu_{R_e}^* = \delta^* = 0.4$ \textit{radians} for all robots. Left: alignment errors $\delta_i(t)$ and variation $\Delta \lambda_\text{min}(t)$. Right: robot trajectories, and body axes ($x_{B_i}$, red, $y_{B_i}$, green, $z_{B_i}$, blue).
    }
    \label{fig: thm}
\end{figure}

\section{Conclusion}
\label{sec: conclusion}

We extended our previous source-seeking framework to 3D by developing a geometric controller on $\text{SO}(3)$, avoiding singularities and ambiguities while ensuring alignment with the desired ascending direction and maintaining non-degenerate swarm deployments. The design preserves one degree of freedom, enabling adaptation to specific missions or platforms. Future work will integrate this controller on fixed-wing robots and combine it with dispersion-based algorithms to strengthen deployment guarantees and exploit swarm flexibility.







\bibliography{bibtex}             

\end{document}